\documentclass[journal]{IEEEtran}

\usepackage{xcolor,soul,framed} 
\usepackage{verbatim}
\colorlet{shadecolor}{yellow}
\usepackage[pdftex]{graphicx}
\graphicspath{{../pdf/}{../jpeg/}}
\DeclareGraphicsExtensions{.pdf,.jpeg,.png}

\usepackage[cmex10]{amsmath}
\usepackage{array}
\usepackage{mdwmath}
\usepackage{mdwtab}
\usepackage{eqparbox}
\usepackage{url}
\usepackage{multirow}
\usepackage[caption=false]{subfig}
\usepackage{graphicx}
\usepackage{multirow}
\usepackage{amsfonts}
\usepackage{cite}
\usepackage{gensymb}

\usepackage{graphicx}

\usepackage{nomencl}
\makenomenclature

\makeatletter
\def\footnoterule{\kern-3\p@
 \hrule \@width 2in \kern 2.6\p@} 
\makeatother

\begin{document}


\title{Supervised Learning based Method for Condition Monitoring of Overhead Line Insulators using Leakage Current Measurement}
  \author{Mile~Mitrovic,
          Dmitry~Titov,
          Klim~Volkhov,
          Irina~Lukicheva,
          Andrey~Kudryavzev,
          Petr~Vorobev,
          Qi~Li,
          Vladimir~Terzija
          
  \thanks{Mile Mitrovic, Dmitry Titov and Klim Volkhov are with Skolkovo Institute of Science and Technology, Moscow, Russia. (email: mile.mitrovic@skoltech.ru)}
  \thanks{Irina Lukicheva is with MIG, Moscow, Russia.}
  \thanks{Petr Vorebov is with Nanyang Technological University, Singapore.}
  \thanks{Qi Li is with National Grid Plc, London, UK.}
  \thanks{Vladimir Terzija is with Newcastle University School of Engineering, Newcastle, UK.}
  }

\maketitle

\begin{abstract}
As a new practical and economical solution to the aging problem of overhead line (OHL) assets, the technical policies of most power grid companies in the world experienced a gradual transition from scheduled preventive maintenance to a risk-based approach in asset management. Even though the accumulation of contamination is predictable within a certain degree, there are currently no effective ways to identify the risk of the insulator flashover in order to plan its replacement. This paper presents a novel machine learning (ML) based method for estimating the flashover probability of the cup-and-pin glass insulator string. The proposed method is based on the Extreme Gradient Boosting (XGBoost) supervised ML model, in which the leakage current (LC) features and applied voltage are used as the inputs. The established model can estimate the critical flashover voltage ($U_{50\%}$) for various designs of OHL insulators with different voltage levels. The proposed method is also able to accurately determine the condition of the insulator strings and instruct asset management engineers to take appropriate actions. 
\end{abstract}



\section{Introduction}

A large percentage of outages on overhead lines (OHLs), especially those in heavily polluted areas, are correlated to insulator flashovers \cite{kordkheili2010determining}. A process of contamination-caused flashover is defined by the insulator type, contaminants, humidity, and the applied voltage \cite{epri2005transmission}. The accumulation of hydrophilic contaminants from roads, cultivated fields, factories, and saltwater bodies on the insulator surface are the most significant factor determining insulator critical flashover voltage ($U_{50\%}$)\footnote{- $U_{50\%}$ is, therefore, the voltage where the probability of flashover of the insulation is 50\%} reduction. This effect is amplified under wet conditions, which leads to more flashover accidents causing line tripping and outages. Therefore, it is necessary to predict the $U_{50\%}$ voltage which indicates the surface condition of the insulator string, and undertake actions accordingly, such as insulator cleaning or replacement. All that can be part of the general procedure for asset management and condition monitoring. In this specific case, analysis is focused on insulator strings at a given voltage level in the power grid.

The technical policy of most electricity transmission owners worldwide is subject to a gradual transition from the traditional method, which is condition monitoring based on a system of scheduled preventive repairs, to a new risk-oriented approach in the company's asset management as a solution to the equipment aging problem. The same approach is applied to OHL insulator management, but even though the accumulation of contamination is predictable, there are currently no effective ways to identify the potential risk of the flashover on the insulator in order to effectively plan its replacement.

The most common method to monitor the aging condition of insulators is visual diagnostics during walking inspections. Visual diagnostics is subjective and only allows obvious damage on the non-glass parts to be identified, with traces left by flashover or very heavily contaminated insulators. There are several methods for insulators’ diagnostics, including diagnostics with modern portable ultraviolet (UV) cameras with partial discharge counting function \cite{kim2011characteristics}, diagnostics with infrared cameras \cite{zhao2016representation}, etc. Diagnostics with UV cameras improve the quality of diagnostics but require further research to improve the data interpretation. It is also crucial to simultaneously capture the meteorological conditions on the site and record the type and number of insulator sheds in the string. Diagnostics with infrared cameras are feasible but require a few conditions to conduct, such as ascent to the tower, cloudy weather without wind or precipitation, and temperature to be above $0$ degrees Celsius. Other diagnostic methods based on revealing equivalent salt deposit density (ESDD) or non-soluble deposit density (NSDD) \cite{cigre1}, which require line outage and dismantling of the line and removing the tested insulator from cross-arm, so they rarely prove their cost-effectiveness.

Direct measurements of some insulator status parameters in situ have become widely deployed globally over the last decade. This is driven by the fast development of sensors and Internet of Things-based technologies, as well as the improved data acquisition and storage capability. 

Some researchers empirically reveal the threshold of a measured parameter which indicates the danger of the flashover, and therefore set up their monitoring system to alert the operator about the potential flashover risk. In \cite{fierro1996line, ramirez1999criteria}, it is concluded, that the insulator should be replaced or cleaned when the leakage current (LC) peaks are greater than 250-450 mA depending on its profile. It should be noted, that the measured level of LC peaks is strongly dependent on the measurement bandwidth, the monitoring methods and the approach in post-processing the obtained results. In the aforementioned research, LC was measured using a full wave rectifier with measurement bandwidth fixed at $1$ kHz and the averaging based on $100$ samples, which are taken over a sampling period of $2$ hours.

Many researchers design monitoring strategies based on the assessment of some insulator state parameters, which are correlated to flashover probability. The assessment relies on selecting measurements or observations in advance. For example, the dependence between the time to flashover on the polluted insulator and the voltage level, the resistance per unit length and the length of the insulator string are used to estimate the flashover probability \cite{ghosh1995estimation}. In \cite{de2020cumulative}, the LC peak is obtained from the meteorological and environmental data, including directional dust deposit conductivity which is difficult to measure directly, using a Random Forest (RF) LC prediction model. Reference \cite{bueno2023inception} focuses on the implementation of a 1D-convolutional neural network (1D-CNN) to predict LC based on environmental data collected during the normal operation of electrical insulators. The approach involves long-term recording to enhance the accuracy of predictions and aims to prevent contamination flashover events. Similarly, in the context of UHF (Ultra-High Frequency) measurements, a Long Short-Term Memory (LSTM) model is employed to assess the temporal evolution of contamination flashover on a polluted insulator in \cite{orellana2023danger}.

In \cite{liu2017self} a reduction in symmetry of self-normalizing multivariate (SNM) maps of LC parameters indicates the discharge propagation and the increasing flashover process stage. Different stages are defined by their proximity to flashover and differ by leakage current. The higher the stage number, the worse the insulator status and the higher the flashover probability is. According to the commonly recognized model \cite{iec1}, the process of the contamination-caused flashover (type A contamination) consists of six stages. In \cite{liu2017self} the authors consider 4 stages, which are easier to distinguish from SNM maps. For insulator strings in service, the stage of the flashover process on a particular sample may change within a few hours or minutes depending on the insulator's moisture content \cite{de2021research}. Consequently, the LC parameters can change dramatically as well. Therefore, the assessment of the instantaneous flashover process stage does not identify the flashover risk accurately when weather conditions are subject to change.

In studies \cite{li2009contamination, ahmad2004assessment,dey2022insulator} LC parameters are used to estimate surface contamination level. The advantage of this approach is the ability to assign one of the site contamination severity classes to infer whether the insulator has to be replaced immediately \cite{iec1, epri2005transmission}. Moreover, it is possible to estimate the insulator critical flashover voltage $U_{50\%}$ based on the contamination severity assessments using empirical equations \cite{kontargyri2007design}. 

In addition to the generally accepted criteria, there are other criteria for making a decision on the action to replace or clean the insulator such as the time-integral of the LC \cite{ghosh2017novel}, standard deviation multi-resolution analysis (STD-MRA) distortion ratio pattern of the LC \cite{chandrasekar2009investigations, douar2010flashover}, etc. Our experience has shown that the most appropriate approach for making the decision on insulator replacement or cleaning is to monitor the $U_{50\%}$ voltage, since the level of the flashover voltage determines the flashover probability. According to the simplified deterministic design approach \cite{epri2005transmission}, statistical withstand voltage equal to the maximum phase-ground voltage $U_{ph}$. To determine the probability of the flashover, it is necessary to determine the proximity between $U_{50\%}$ and $U_{ph}$. Determining $U_{50\%}$ and its standard deviation can be done through laboratory experiments since it is almost impossible to reproduce the flashover in situ. The insulator condition monitoring based on surface conductivity \cite{gutman2005application}, salt deposit density (SDD) \cite{salem2021prediction}, LC integral parameters \cite{zhao2013flashover}, frequency characteristics of LC waveforms \cite{suda2001frequency}, etc. can also be created. All these measured parameters obtained in situ can be used to decide whether to take actions to prevent insulator flashovers, such as insulator cleaning, washing, or replacement \cite{de2020cumulative}.

Nevertheless, the defined relationships between measured parameters and $U_{50\%}$ voltage given in \cite{gutman2005application,salem2021prediction,zhao2013flashover,suda2001frequency} can be used only for the same string configuration and types of insulator which are under the same voltage level as applied in the research laboratories. This severely limited the applicability of the developed mathematical models in practice.

To improve the prediction, this paper proposed a novel method for condition monitoring of OHL insulator strings based on the Extreme Gradient Boosting (XGBoost) supervised machine learning (ML) model. The method allows to input the weather conditions such as wet or dry and to classify the status of the insulator string. The classification is based on the following three states: 1) operational state, 2) hazardous state, and 3) extremely hazardous state. The method is based on measurements of LC parameters in situ and can be used for any number of cap-and-pin glass insulators in a string and for any OHL voltage level.

\section{ML-based Insulators' Condition Monitoring Method}\label{sec:2} 

\begin{figure}[ht]
  \begin{center}
  \includegraphics[width=3.2in]{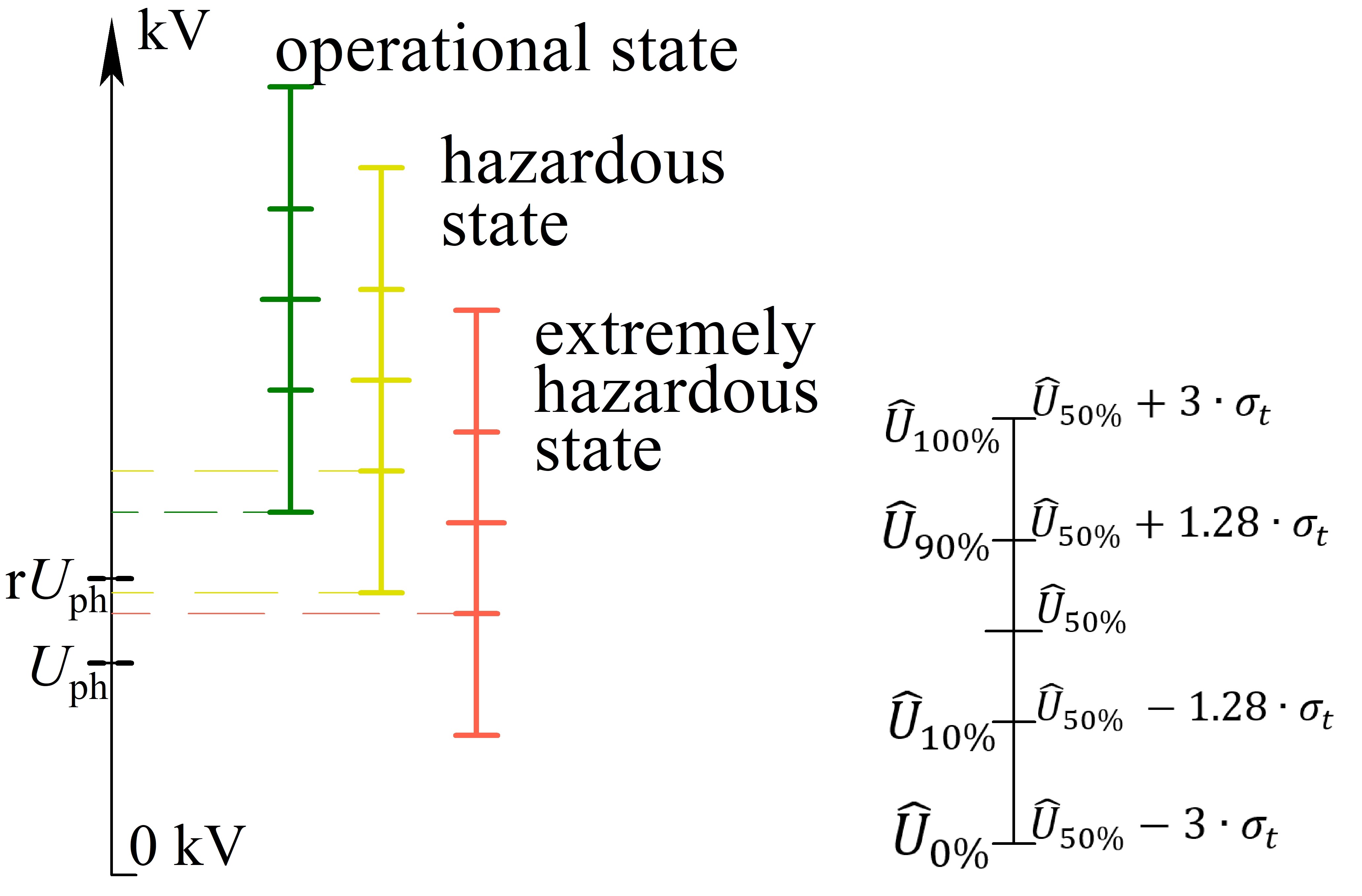}
  \caption{Conditions for the classification of insulator string states; $U_{xx\%}$ is the estimated voltage where the probability of flashover of the insulator is $xx\%$.}
  \label{fig:1}
  \end{center}
  \vspace{-3mm}
\end{figure}

Classification of insulator string states is based on comparison of the \textit{estimated} critical flashover voltage $\hat{U}_{50\%}$ with OHL phase-to-ground voltage $U_{ph}$ (Fig.~\ref{fig:1}). It is well-known that the degree of surface humidity strongly affects the $U_{50\%}$ voltage. When the surface is under wet conditions, the $U_{50\%}$ voltage drops down and approaches to the  $U_{ph}$ voltage. This is additionally manifested through a larger contamination layer.

The right part of Fig.~\ref{fig:1} presented the level of $U_{50\%}$ voltage and its deviation for three different insulator strings, obtained under the same humidity degree. The difference of $U_{50\%}$ voltage for the insulator strings is defined by the difference in contaminations. In Fig.~\ref{fig:1}, $\sigma_t$ is the total standard deviation of $\hat{U}_{50\%}$, kV:
\begin{equation}\label{eq:1}
    \sigma_t = \sigma + \hat{\sigma}_m
\end{equation}
where $\hat{\sigma}_m$ is the standard deviation of the $\hat{U}_{50\%}$, kV; $\sigma$ is the standard deviation of the actual $U_{50\%}$ voltage determined from the test results, kV. In any case, the contamination-caused flashover of an insulator string is considered to be a practically rare event if:
\begin{equation}\label{eq:2}
    r\cdot U_{ph} < \hat{U}_{50\%} - 3\cdot \sigma_t
\end{equation}
where $U_{ph}$ is the OHL phase-to-ground voltage, kV; and parameter $r$ is a safety factor. The safety factor is chosen taking into account the ampacity of the line and the number of strings. For OHLs shorter than $100$ $km$ we can use $r=1.6$ (Fig. 4.8-14 in \cite{epri2005transmission}).

The fulfilling condition in (\ref{eq:2}) corresponds to the operational state of the insulator string (green line in Fig.~\ref{fig:1}). The contamination-caused flashover of an insulator string is considered to be a low probability (less than $10\%$) event if:
\begin{equation}\label{eq:3}
    \hat{U}_{50\%} - 3\cdot \sigma_t \leq r\cdot U_{ph} < \hat{U}_{50\%} - 1.28\cdot \sigma_t
\end{equation}
The fulfilling condition in (\ref{eq:3}) corresponds to the hazardous state of the insulator string (yellow line in Fig.~\ref{fig:1}). The extremely hazardous state of the insulator string (red line in Fig.~\ref{fig:1}) is defined as:
\begin{equation}\label{eq:4}
    r\cdot U_{ph} \geq \hat{U}_{50\%} - 1.28\cdot \sigma_t
\end{equation}

To be able to use the developed mathematical model of the insulator state for any number of cap-and-pin glass insulators in a string and for any voltage level of an OHL, $U_{ph}$ is expressed through the \textit{percentage} of $\hat{U}_{50\%}$ ($\%\hat{U}_{50\%}$) as an output parameter of the model. In this case, $\hat{U}_{50\%}$ is defined as following:
\begin{equation}\label{eq:5}
    \hat{U}_{50\%} = \frac{100 \cdot U_{ph}}{\%\hat{U}_{50\%}}
\end{equation}
where $\%\hat{U}_{50\%}$ is a value in $\%$, which means the degree of percent $U_{ph}$ is from $\hat{U}_{50\%}$. Also, it is necessary to determine $\hat{\sigma}_m$ of each new estimated data according to the testing data errors:
\begin{equation}\label{eq:6}
    \hat{\sigma}_{m} = \frac{\%\hat{\sigma}_m \cdot \hat{U}_{50\%}}{100}
\end{equation}
where $\%\hat{\sigma}_m$ is a root mean square error (RMSE) for $\%\hat{U}_{50\%}$ in the developed mathematical model, $\%$.

\begin{figure}[ht]
  \begin{center}
  \includegraphics[width=3.2in]{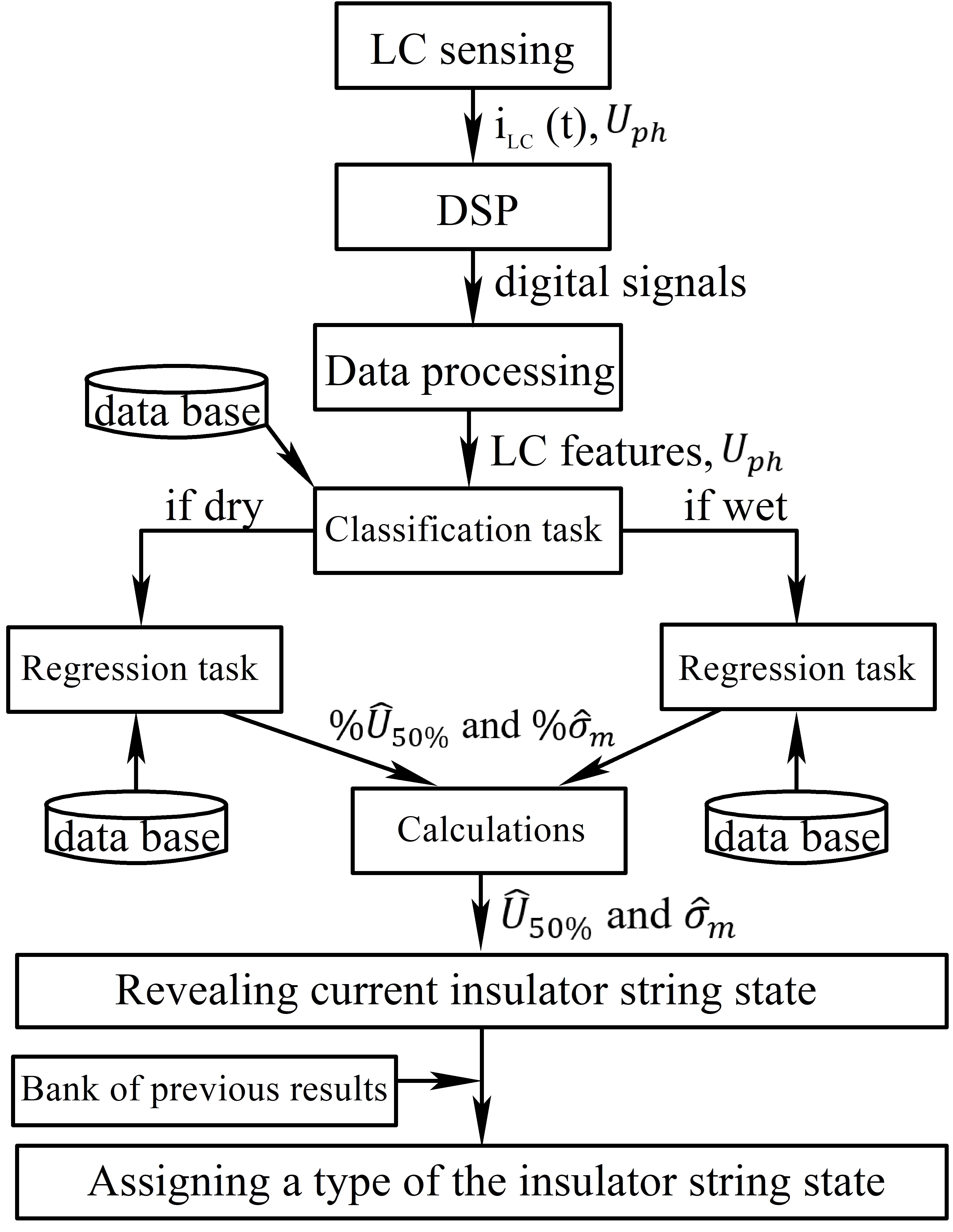}
  \caption{Schematic diagram of assigning a type of insulator string state based on LC measurements.}
  \label{fig:2}
  \end{center}
\end{figure}

The procedure to classify the insulator string state can be divided into the following steps (Fig.~\ref{fig:2}):
\begin{itemize}
\item[--] obtaining the LC waveform from the diagnosed insulator string;
\item[--] digital signal processing (DSP);
\item[--] extracting LC features from the waveform, which mostly correlates with $U_{50\%}$ voltage under wet and dry conditions;
\item[--] determination of the insulator string surface condition (dry or wet);
\item[--] revealing $\%\hat{U}_{50\%}$ and $\%\hat{\sigma}_m$;
\item[--] calculating $\hat{U}_{50\%}$ and $\hat{\sigma}_m$ based on (\ref{eq:5}-\ref{eq:6});
\item[--] checking the conditions (\ref{eq:2}-\ref{eq:4}) and revealing the current insulator string state: operational state, hazardous state, or extremely hazardous state.
\end{itemize}

It is also found that the instantaneous value of $\hat{U}_{50\%}$ is not decisive. The worst case of the insulator string during the last 2-3 months should be considered to determine the state, even if the last results showed a better state. This paper is devoted to determine LC features that mostly correlate with $U_{50\%}$ voltage under wet and dry conditions, as well as creating the database for solving the classification task (whether an insulator string is under wet or dry conditions) and regression task separately for wet and dry insulators (what the level of  $\%\hat{U}_{50\%}$ and $\%\hat{\sigma}_m$ is). 

The advantage of the proposed method is that two ML models to estimate $U_{50\%}$ voltage for insulators under wet and dry conditions are developed separately. This effectively reduces the error of estimation and makes the method more robust. In addition, an ML-based data-driven method is created to estimate the insulator state.

The rest of this paper is organized as follows: Section \ref{sec:3} describes the laboratory setup, data preparation and feature extraction of the LC. The description of the XGBoost model for $U_{50\%}$ voltage estimation is given in Section~\ref{sec:4}. Section \ref{sec:5} presents the validation results of various ML models tested for the proposed method. Section~\ref{sec:6} presents a summary of the paper and possible extensions.

\section{Laboratory setup and data collection}\label{sec:3} 

To estimate the flashover probability model according to the described steps in Section~\ref{sec:2}, data on LC under dry and wet conditions with various contamination levels and $U_{50\%}$ voltage are required. The labeled data is obtained through high-voltage experiments.

\subsection{Laboratory Facility}
\begin{figure}[ht]
  \begin{center}
  \includegraphics[width=3.2in]{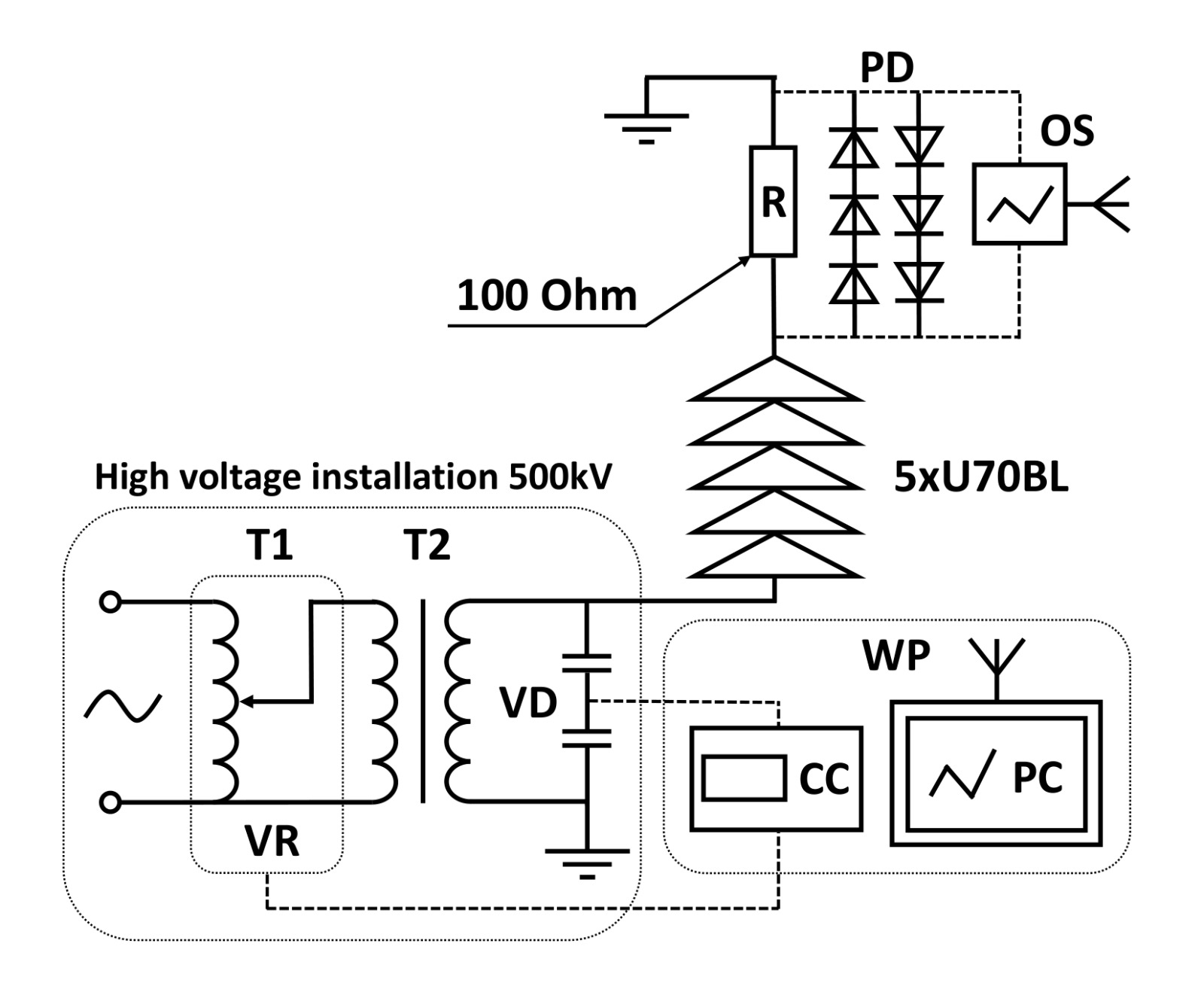}
  \caption{Single-line circuit representing the experimental setup. \newline \textbf{PD} is the protective diodes, \textbf{OS} is the Wi-Fi oscilloscope, \textbf{R} is the non-inductive resistor, \textbf{T1} is the regulating transformer, \textbf{T2} is the power transformer, \textbf{VR} is the voltage regulation block, \textbf{VD} is the voltage divider, \textbf{WP} is the operator's workplace, \textbf{CC} is the control console, \textbf{PC} is the computer with Wi-Fi.}
  \label{fig:3}
  \end{center}
\end{figure}

The experiments were carried out in the High Voltage Laboratory of the Yuzhnouralsky Insulators and Fittings Plant. The principal electrical diagram is shown in Fig.~\ref{fig:3}.

\begin{figure}[ht]
  \begin{center}
  \includegraphics[width=3.2in]{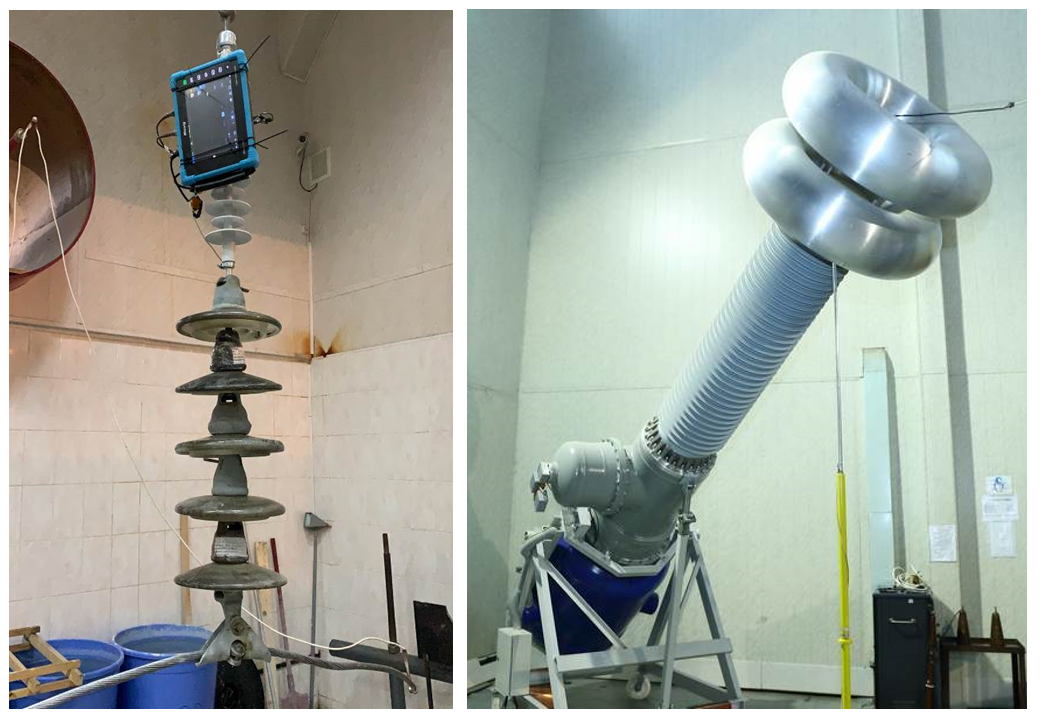}
  \caption{left - high voltage installation UIV-500 (with voltage level up to 500 kV, load up to 60 kVA, and sustained short circuit current 0.3 A); right - insulator string under testing. }
  \label{fig:4}
  \end{center}
  \vspace{-5mm}
\end{figure}

The installed transformer UIV-$500$ was used as the HV supply (Fig.~\ref{fig:4} - left). Voltage regulation was conducted using the control console, which displays the applied voltage value in kV. In order to comply with the requirements listed in the IEC $60507$ on the energized voltage of at least $700$ V per meter for the overall creepage distance, data obtained at a voltage below $1.5$ kV (which was no more than $2\%$ $U_{50\%}$ voltage) were not taken into account.

According to the table in Appendix A of IEC 60407, the highest level of specific creepage distance $L_s$ was up to $8$ mm/kV (close to discharge voltage), which is equivalent to approximately $0.12$ A of the highest leakage current pulse amplitude $I_h$. The sustained short circuit current for the testing transformer UIV-$500$ is lower than the required level in IEC $60507$, but the developed model used $I_h$ obtained on the voltage level close to phase voltage where $I_h$ $\ll$ $0.3$/$11$ $A$. Therefore, the testing set was appropriate for the experiment’s purpose. 

Voltage was applied to the test sample from the bottom side. In order to capture the waveform for LC, an oscilloscope \textit{Micsig TO$1104$ plus} was used. To install the oscilloscope and create a break in the electrical circuit, a core polymer insulator LK-$70$/$35$ was used (Fig.~\ref{fig:4} - left). 

The measurements were taken on the grounded side of the strings through an AN-$25$ resistor with a nominal value of $100$ Ohm synchronously on two channels: 
\begin{enumerate}
\item with a division setting of $0.1$ V to measure the amplitude of the harmonic component in the range of up to $5$ mA; 
\item with a division setting of $1$ V to measure the amplitude of the pulse component in the range up to $0.5$ A.
\end{enumerate}

PC Software 'TO $1000$ Series' was used to control the oscilloscope remotely and save the measurement results. In order to prevent damage to the oscilloscope caused by flashover, a protection circuit consisting of counter-switched zener diodes $1N5359B$ with a reverse current of no more than $0.5$ $\mu$A was used.

\subsection{Test Samples}
The behavioral patterns of amplitudes rising are different for artiﬁcially and naturally polluted insulators \cite{swift2000polluted}. For artiﬁcially polluted insulators, the ﬂashover is accompanied by less current amplitude and amplitude rising is more straightforward. It can be explained by its better wettability and uniformity of accumulated contamination. For this reason, it was decided to avoid using artiﬁcially polluted insulators for the experiments.

\begin{figure}[ht]
  \begin{center}
  \includegraphics[width=3.5in]{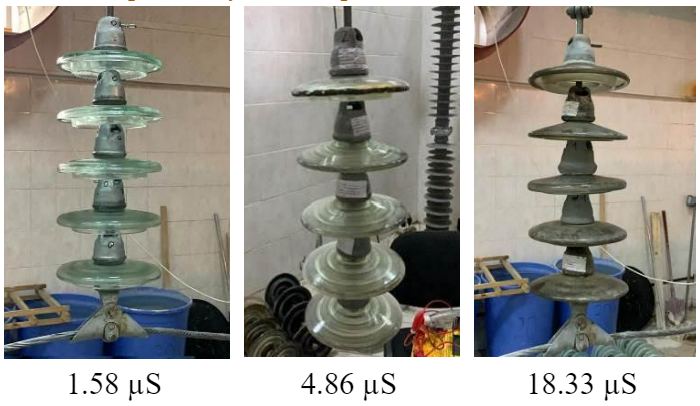}
  \caption{Insulator strings without (1.58 $\mu S$) and with natural contamination (4.86 $\mu S$ and 18.33 $\mu S$). 
  Surface conductance was measured under wet conditions by high voltage insulation tester Sew 2804 IN.
}
  \label{fig: ins1}
  \end{center}
  \vspace{-5mm}
\end{figure}

Three glass insulator strings assembled from the most common cup-and-pin insulators $U70BL$ with different surface conductivities were used as test samples (Fig.~\ref{fig: ins1}). Two of them were in service for the period of 32 and 44 years respectively before decommissioning, which represents naturally contaminated conditions. 
The third insulator string was a clean sample without any contamination for comparison reasons.

\subsection{Test procedures}
\subsubsection{Preparation of the test sample} the wet layer of the insulator surface was created by precooling the insulator string to form a uniform layer of dew on its surface. For this purpose, the strings were placed into a climate chamber with an internal temperature of about $2^{\degree}C$ for $2$ hours. After taking it out from the climate chamber, the surface of the glass was covered by a layer of dew uniformly within $10$-$15$ min. This procedure kept moisture on the surface even after several testing cycles and ensured the reproducibility of the experiment. 
\subsubsection{$U_{50\%}$ voltage revealing} after connecting the wet insulator string into the setup, the applied voltage was raised up steadily to flashover level according to IEC $60507$. The test was repeated $10$ times with an interval of about $1$ min. The fixed levels of  flashover voltage $U_f$ were used to calculate the average discharge voltage $\bar{U}_{av}$:
\begin{equation}\label{eq:7}
    \bar{U}_{av} = \frac{1}{N} \sum_{i=1}^{N} U_{f_i}
\end{equation}
and estimated value of the standard deviation $\sigma$ as:
\begin{equation}\label{eq:8}
    \sigma = \sqrt{\frac{1}{N-1}\sum_{i=1}^{N}(U_{f_i} - \bar{U}_{av})^2}
\end{equation}
where N is the number of flashover tests. The standard deviation $\sigma$ of measurement results from $U_{50\%}$ voltage averaged $14$ kV for all strings.

Based on the obtained $\bar{U}_{av}$ and $\sigma$, the lowest value of the average flashover voltage $\bar{U}^{'}_{av}$ and the highest value in relative units of standard deviation $\sigma^{'}$ were determined using formulas valid for a confidence probability of $90$\%: 
\begin{equation}\label{eq:9}
    \bar{U}^{'}_{av} = \bar{U}_{av} - 0.572\sigma
\end{equation}
\begin{equation}\label{eq:10}
    \sigma^{'} = \frac{1.64\sigma}{\bar{U}_{av}}
\end{equation}
Sustained voltage $U_{50\%}$ voltage with a given probability of $90\%$ is defined by IEC $60061$ as:
\begin{equation}\label{eq:11}
    U_{50\%} = \bar{U}^{'}_{av}(1-1.3\sigma)
\end{equation}

\subsubsection{Leakage current measurements} to determine the relationship between $\%U_{50\%}$ (Eq.~\ref{eq:11}) and LC features the voltage was raised up steadily to $95$-$100\%$ of  $U_{50\%}$ voltage. The rate of voltage increment was chosen such that for $10$ periods of $50$ Hz the voltage increment did not exceed $0.5$-$1\%$ of $U_{50\%}$ voltage. The leakage current was measured continuously throughout the test. 

The insulator strings were tested in accordance with the above-mentioned methods under dry and wet conditions. The obtained results were used for a signal feature extraction.

\subsection{Signal feature extraction}

\begin{figure}[ht]
 \centering
 \subfloat[\centering Actual and filtered LC waveform;]{{\includegraphics[width=0.8\linewidth]{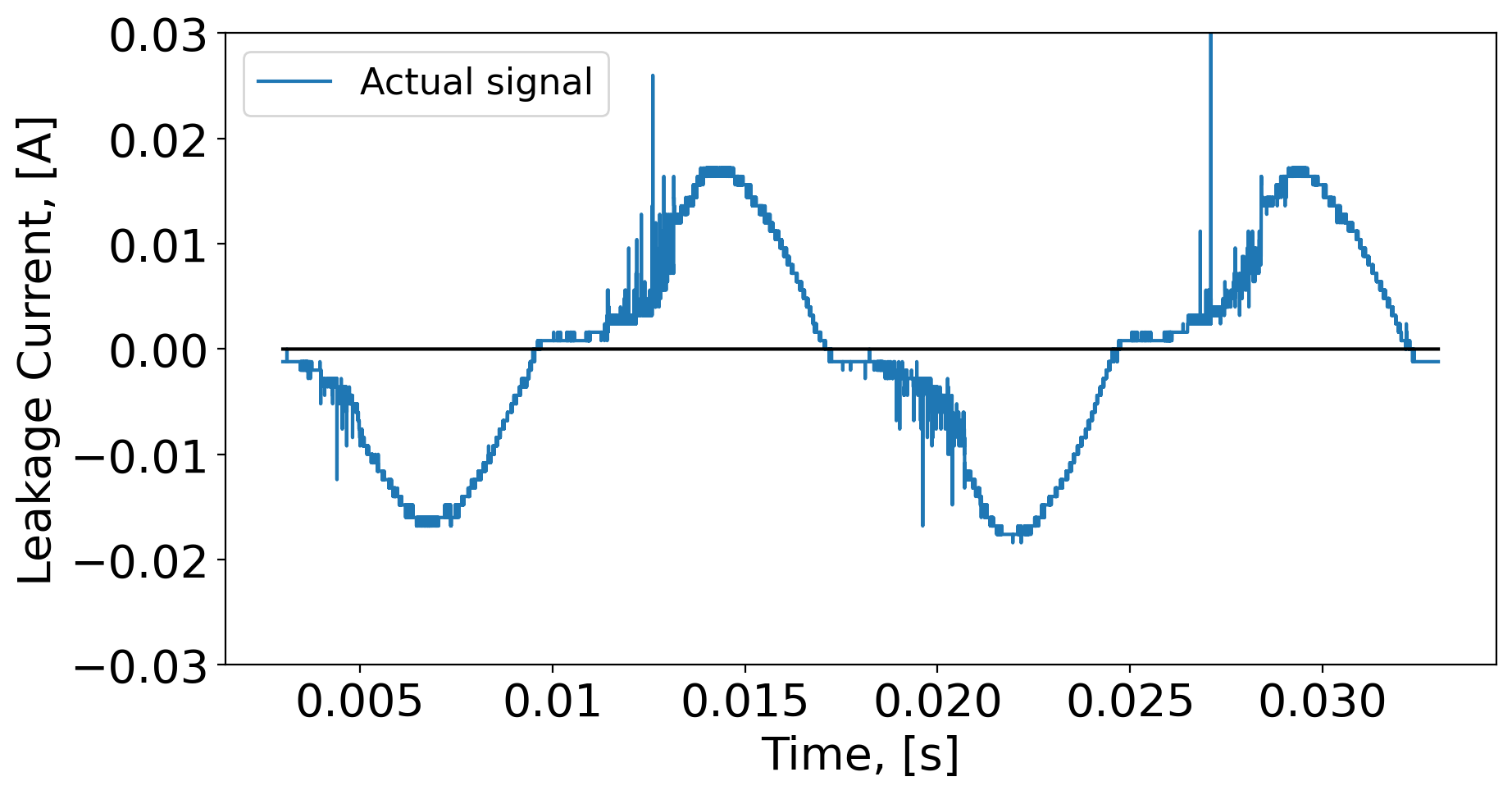} }
 \label{fig:LC_a}}%
 \qquad
 \subfloat[\centering LC pulses;]{{\includegraphics[width=0.8\linewidth]{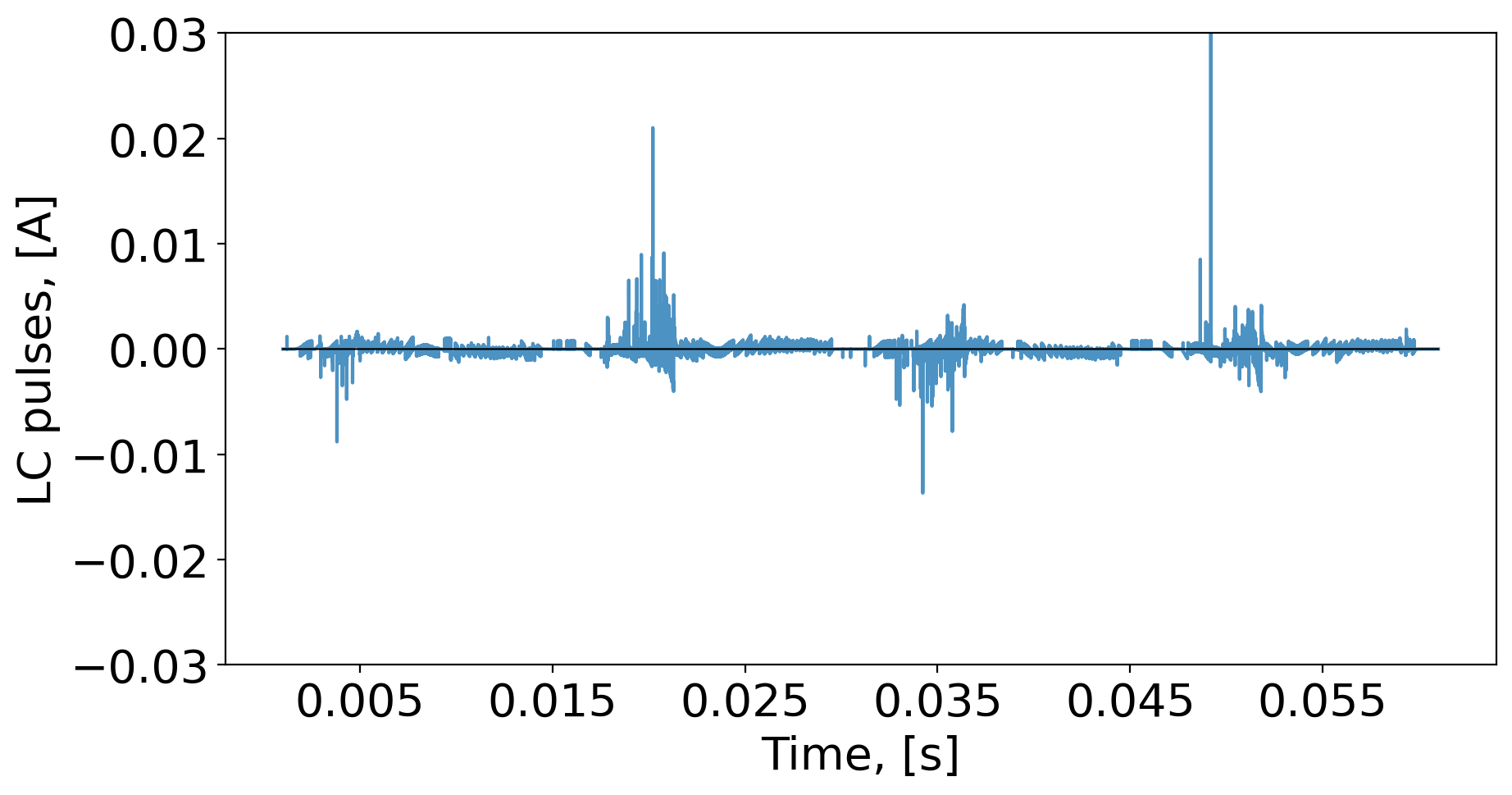} } \label{fig:LC_b}}%
 \qquad
 \subfloat[\centering LC frequency spectrum; ]{{\includegraphics[width=0.8\linewidth]{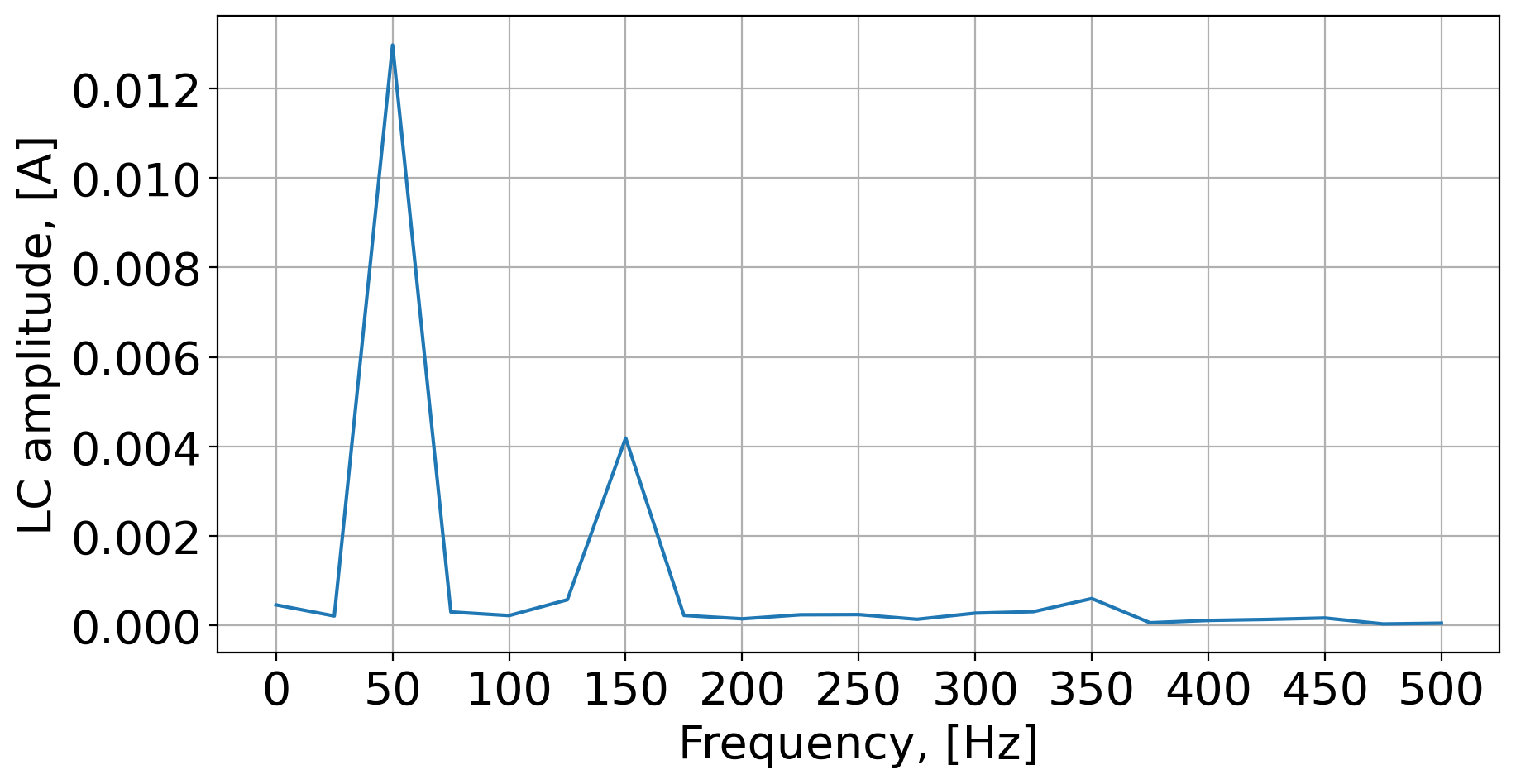} }
 \label{fig:LC_c}}%
 \caption{LC waveform, pulses and frequency spectrum of the contaminated insulator in wet condition.}%
 \label{fig:LC}%
 \vspace{-3 mm}
\end{figure}

\begin{figure}[ht]
  \begin{center}
  \subfloat[\centering wet conditions;]{{\includegraphics[width=1.7in]{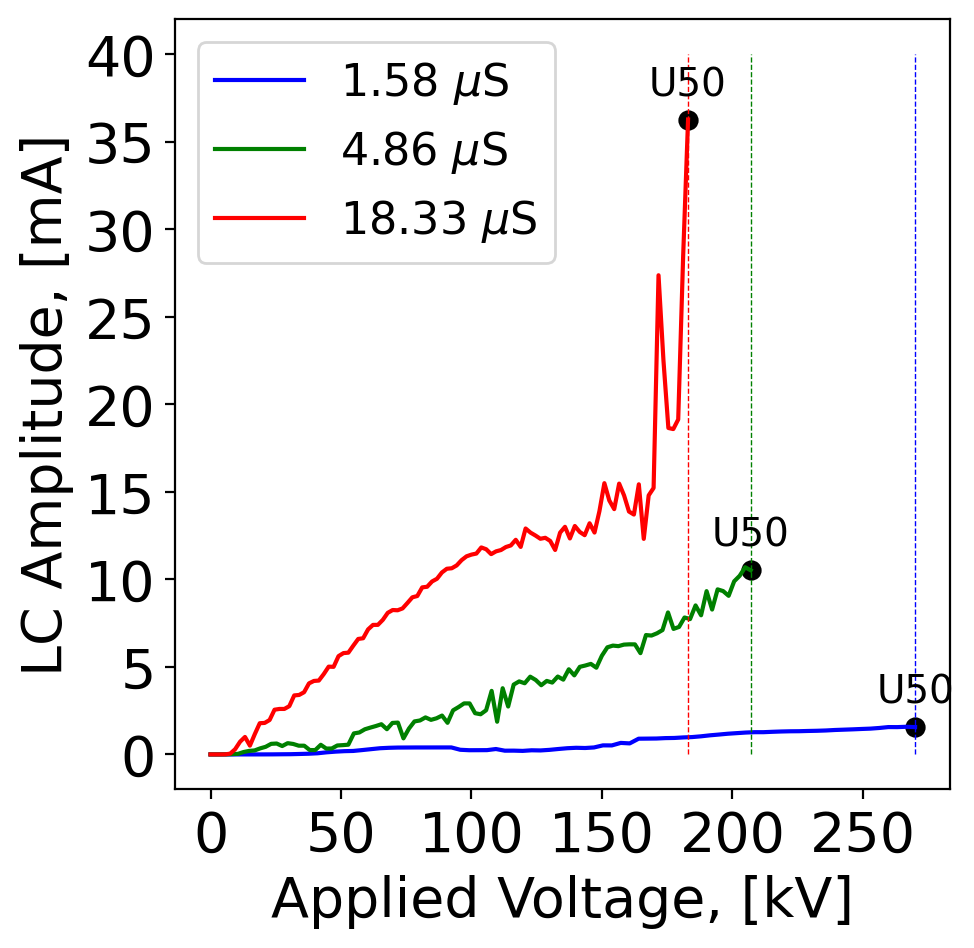} }
 \label{fig:7a}}%
\subfloat[\centering dry conditions;]{{\includegraphics[width=1.7in]{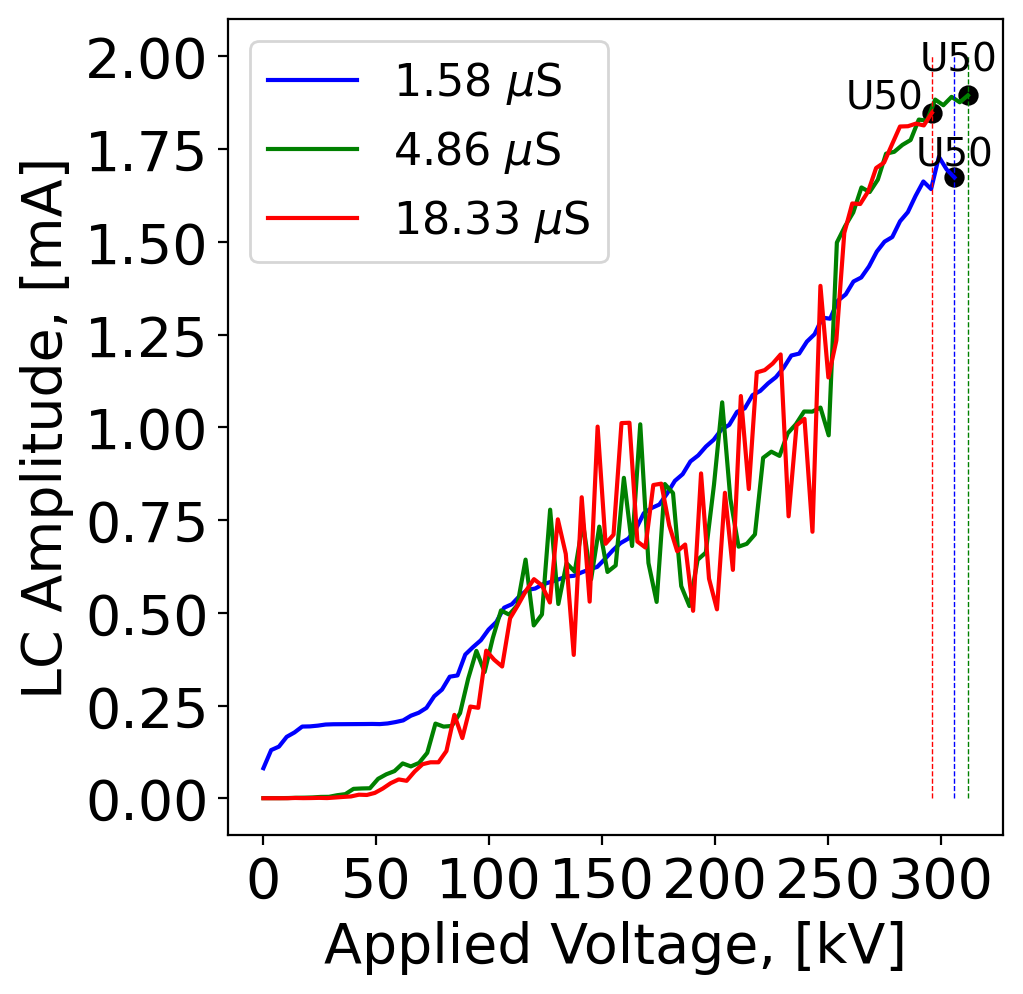} }
 \label{fig:7b}}%
  \caption{Dependence of the insulators LC amplitude with various contamination levels in the dry and wet conditions on the applied voltage. (The highest voltage points represent $U_{50\%}$ voltage; the contamination level is described with layer conductance in micro-Siemens (µS))}
  \label{fig:7}
  \end{center}
  \vspace{-5mm}
\end{figure}

Features of the LC waveform are extracted to prepare for $U_{50\%}$ voltage estimation based on ML. In this paper, the LC features in both the time and frequency domains are considered to cover all possible relationships between LC features and $U_{50\%}$ voltage. A typical LC waveform of the contaminated insulator under the wet condition is shown in Fig.~\ref{fig:LC_a} (obtained during the experiments). The actual LC waveform is accompanied by pulses that characterize the development of discharge. Additionally, contamination and wet conditions are associated with large pulses (Fig.~\ref{fig:LC_b}) and significant signal distortion. The amplitudes of the harmonic components are calculated by transforming the LC waveform into the frequency domain using the Fast Fourier Transform (Fig.~\ref{fig:LC_c}). Fig.~\ref{fig:7} shows the results for the amplitudes of the harmonic components (in $mA$), for three insulating samples.

A total of $63$ LC features were extracted from the experimental results. Two low-pass filters are employed in each case to smooth the actual signal and remove background noise, using moving average (MA) and exponential smoothing (ES) methods. The filtered signals are used to extract the fundamental waveform (first harmonic) at $50$~Hz. The difference between the actual and the fundamental waveform is used to highlight features in the time domain. Accordingly, features of the statistical parameters are extracted for samples received at differences between actual waveform and fundamental waveform derived from both filtered MA and ES waveforms. These features are: mean and standard deviation; min, max, absolute maximum between min and max values, and various functions of absolute max such as square, square root, natural log, common log, exponential function, inverse proportion, and power of $10$, $25$th, $50$th, and $75$th percentile, mean and amount of all extracted pulses. The amplitude of the fundamental waveform obtained from the MA-filtering is considered to be the LC feature, as well as various functions of this amplitude such as square, square root, natural log, common log, exponential function, inverse proportion and power of $10$. The features of the LC pulses were divided into $12$ ranges in $mA$ and $10$ ranges as a percentage of the amplitude for the fundamental waveform. These features are the number of pulses in the ranges: less than $0.2$ mA, $0.2$ to $0.5$ mA, $0.5$ to $1$ mA, $1$ to $2$ mA, $2$ to $3$ mA, $3$ to $4$ mA, $4$ to $5$ mA, $5$ to $7.5$ mA, $7.5$ to $10$ mA, $10$ to $15$ mA, $15$ to $20$ mA, more than $20$ mA, less than $50\%$ of fundamental amplitude, $50\%$ to $100\%$,  $100\%$ to $200\%$,  $200\%$ to $300\%$, $300\%$ to $400\%$, $400\%$ to $500\%$, more than $500\%$ of fundamental amplitude respectively. The features derived from the frequency domain are the amplitudes from the 1st to the $10$th harmonic. 
In addition to the LC features, we consider the applied voltage as an input feature. Thus, the proposed method based on ML can be successfully applied for the $U_{50\%}$ voltage estimation of an insulator at different voltage levels.

\section{Machine Learning for $U_{50\%}$ voltage Estimation}\label{sec:4} 

The proposed method in Section~\ref{sec:2} is designed to estimate the insulator's $U_{50\%}$ voltage using the extracted features of LC for any voltage level. This ML-based method consists of two parts: classification task and regression task (Fig.~\ref{fig:2}). In the classification task, the condition of the insulator is identified by using the XGBoost model for classification. Once the wet or dry condition is identified, the method estimates $\%U_{50\%}$ and $\%\sigma_{m}$ in the regression task by using the XGBoost model for regression. 

The experiments of the cap and pin glass insulators have shown that amplitudes of the LC signals are significantly higher under wet conditions. Accordingly, the probability of the insulator flashover is higher under wet conditions. Therefore, LC features and their dependencies on insulator $U_{50\%}$ voltage data behave differently in wet and dry conditions. This fact can significantly affect the performance of the ML model. To address this, it is proposed to identify the wet or dry condition first, and then train separate regression models for each condition. Therefore, a more robust method is created which reduces errors in flashover estimation..

For each XGBoost model, a feature selection technique is applied to improve the ML process and model performance. Thus, to ensure that, the most relevant information is extracted from the input data to create ML models with better accuracy. As a feature selection technique, the Maximum Relevance Minimum Redundancy approach is employed.

This investigation reviews and compares other ML algorithms such as Random Forest (RF), K-Near Neighbors (KNN), Linear and Logistic Regression (LiR, LoR). Algorithms such as neural networks and support vector machines were also considered, but the poor results obtained demonstrated that they are not suitable candidates for this specific application.

\subsection{XGBoost model}
XGBoost or Extreme Gradient Boosting (XGB) is a supervised tree-based algorithm for classification and regression tasks. XGBoost presents an implementation of the gradient boosting (GB) decision trees that uses more accurate approximations~\cite{chen2016xgboost}. GB utilizes the individual weak decision tree and iteratively adds new trees to further minimize the objective function. This process continues until the specified number of boosting iterations is reached. Unlike GB, XGBoost uses the second-order gradients of the objective function which contribute to a better prediction on the direction of the objective function minimum. In addition, XGBoost employs regularization techniques such as lasso regression (L1) and ridge regression (L2) which reduce the model complexity and chances for overfitting. XGBoost improves overall generalization and computational speed.

Given data set with $m$ samples and $n$ features: $\{(x_i, y_i)_{i=1}^{m}, ~x_i \in \mathbb{R}^{n}, ~y_i \in \mathbb{R} \}$, where $x_i$ is the $i$-th input sample vector of \textit{LC} features and \textit{applied voltage}; $y_i$ is the corresponding output. Depending on the method whether it is a classification task or regression task, $y_i$ is the condition of the insulator or $\% U_{50\%}$ respectively. The XGBoost model can be expressed as: 
\begin{equation}
    \hat{y}_i = \sum_{k=1}^{K} f_k(x_i)
\end{equation} 
where $f_k$ is a tree at the $k$-th 
instance. A new tree $f_t$ is added iteratively to minimize the regularized objective function as:
\begin{equation}
    \mathcal{L}(t) = \sum_{i=1}^{m} l(y_{i},\hat{y}^{t-1}_i + f_t(x_i)) + \Omega(f_t)
\end{equation}
where $\mathcal{L}$ is the specified objective function (logistic function for classification and squared function for regression), $y_i$ is ground truth, $\hat{y}_i^{t-1} + f_t(x_i)$ is the prediction at the $t-1$ iteration, and $\Omega$ is a regularization term. The regularization term penalizes the complexity of the model as:
\begin{equation}
    \Omega(f_t)  = \gamma T + \frac{1}{2}\lambda
    \sum_{j=1}^{T} w^2_j
\end{equation}
where $T$ is the number of leaves in the tree, $w_j$ is weight of the corresponding leaf, $\gamma$ and $\lambda$ are the L1 and L2 regularization parameters, respectively. For details, we refer the reader to~\cite{chen2016xgboost}.

\subsection{Maximum Relevance Minimum Redundancy}
Feature selection is one of the crucial techniques in the ML pipeline, which targets the most important input features, to make ML models better in prediction. Maximum Relevance Minimum Redundancy (MRMR) is a feature selection approach that tends to select input features with a high correlation with the output (relevance) and a low correlation between themselves (redundancy)~\cite{asefi2022power, radovic2017minimum, asefi2023detection}. In this paper, relevance is calculated by using mutual information due to nonlinear dependency between input features and output~\cite{smith2015mutual}. However, redundancy is calculated by using Spearman’s rank-based correlation because of the processing of non-normality data~\cite{gauthier2001detecting}. 

MRMR works iteratively, where the best feature $x_f$ with maximum relevance and minimum redundancy is selected at each iteration $i$. The score for each feature $x_f$ at each iteration $i$ is computed as following \cite{zhao2019maximum}:
\begin{equation}
    score_i(x_f) = \frac{relevance(x_f|y)}{redundancy(x_f| {x_f}_{selected ~ until ~ i-1})}
\end{equation}
where the best feature $x_f$ at iteration $i$ is the one having the highest score.

\subsection{Evaluation metrics}
The F1 score is used to evaluate the accuracy of the classification task. F1 score represents harmonic mean of the precision $Pr$ and recall $Re$ as:
\begin{equation}
    F1 = 2 \cdot \frac{Pr \cdot Re}{Pr + Re}
    \label{eq: F1_evalMetric}
\end{equation}
where:
\begin{equation*}
    Pr = \frac{TP}{TP+FP}, ~~ Re = \frac{TP}{TP+FN}
\end{equation*}
$TP$ (True Positive) is the number of correctly identified wet conditions; $FP$ (False Positive) is the number of dry conditions identified as wet; $FN$ (False Negative) is the number of wet conditions identified as dry.

Root mean square error (RMSE) is used to evaluate the $U_{50\%}$ voltage error estimation for both regression models as: 
\begin{equation}\label{eq:rms}
     RMSE = \sqrt{\frac{1}{m_*} \sum_{i=1}^{m_*} (\%U_{50\%, i} - \%\hat{U}_{50\%, i})^2}
\end{equation}
where $m_*$ is the number of unseen data and $\%\hat{U}_{50\%, i}$ is the estimated percentage of the $U_{50\%, i}$.

\section{Results}\label{sec:5} 
The results presented in this paper are computed on an Intel Core i7-11370H CPU @ 3.30GHz and 16GB of RAM. The data processing and model training was performed in a Python environment. Both classification and regression XGBoost model hyperparameters are optimized using Bayesian optimization with Gaussian Processes as a surrogate probability model of the objective function \cite{snoek2012practical, asefi2023anomaly}. The dataset was randomly split into $80\%$ of available data for models’ training and the rest $20\%$ for models’ validation. The random split seed has been fixed at \textit{random state = 0} to ensure reproducible splitting of the data.

\subsection {Validation of the classification task of the proposed method}

The F1 score metric is used to validate ML models on an unseen dataset in the classification task of the proposed method. The results of the F1 score metric of various ML models for classification are shown in Table~\ref{table:first_stage}. The experiments were performed for a number of different best features determined using the MRMR approach. The results showed that the highest F1 score is achieved with the proposed XGBoost model for the top $20$ features. 

\begin{table}[ht]
\centering
  \caption{F1 score of the testing classification models for various numbers of selected best features (the classification task).}
 \begin{tabular}{l|l|l|l|l|l|l}
 \hline\hline
  Num. of & LoR & KNN & SVM & NN & RF & XGB \\
  features & [$\%$] & [$\%$]& [$\%$]& [$\%$]& [$\%$]& [$\%$]\\
 \hline
 1& $70.99$ & $78.69$ & $74.89$ & $44.36$ & $73.99$ & $77.77$  \\ \hline
 5& $71.59$ & $87.03$ & $78.70$ & $71.29$ & $86.11$ & $86.72$ \\ \hline
 10 & $72.18$ & $83.33$ & $80.54$ & $73.93$ & $86.11$ & $86.10$ \\ \hline
 15 & $79.52$ & $86.10$ & $82.41$ & $79.51$ & $87.03$ & $87.02$ \\ \hline
 20 & $79.52$ & $85.19$ & $83.32$ & $78.48$ & $87.03$ & $\textbf{95.37}$ \\ \hline
 30 & $88.73$ & $85.16$ & $89.77$ & $86.10$ & $94.44$ & $94.44$ \\ \hline
 40 & $89.80$ & $89.79$ & $90.71$ & $88.85$ & $94.08$ & $93.52$ \\ \hline
 64 & $90.71$ & $88.87$ & $34.14$ & $90.73$  & $93.52$ & $93.12$ \\ \hline\hline
 \end{tabular}
 \label{table:first_stage}
\end{table}

The considered hyperparameters and their optimal values of the XGBoost classification model for the top $20$ input features are presented in Table~\ref{table:first_stage_optm}.

\begin{table}[ht]
\centering
  \caption{Optimal hyperparameters of XGB for classification.}
 \begin{tabular}{l|l|l|l}
 \hline\hline
  Model & Hyperparameters & Search range & Optimal value \\
 \hline
\multirow{4}{*}{XGB} 
                    & n-estimators & [$50-1000$] & $422$   \\\cline{2-4}
                    & max depth & [$3-15$] & $4$  \\\cline{2-4}
                    & learning rate & [$0.001-1$] & $0.157$ \\\cline{2-4}
                    & subsample & [$0.5-1$] & $0.837$ \\\cline{2-4}
                    & colsample by tree & [$0.5-1$] & $0.603$ \\ \hline\hline
                
 \end{tabular}
 \label{table:first_stage_optm}
\end{table}

\begin{figure}[ht]
  \begin{center}
  \includegraphics[width=9.cm]{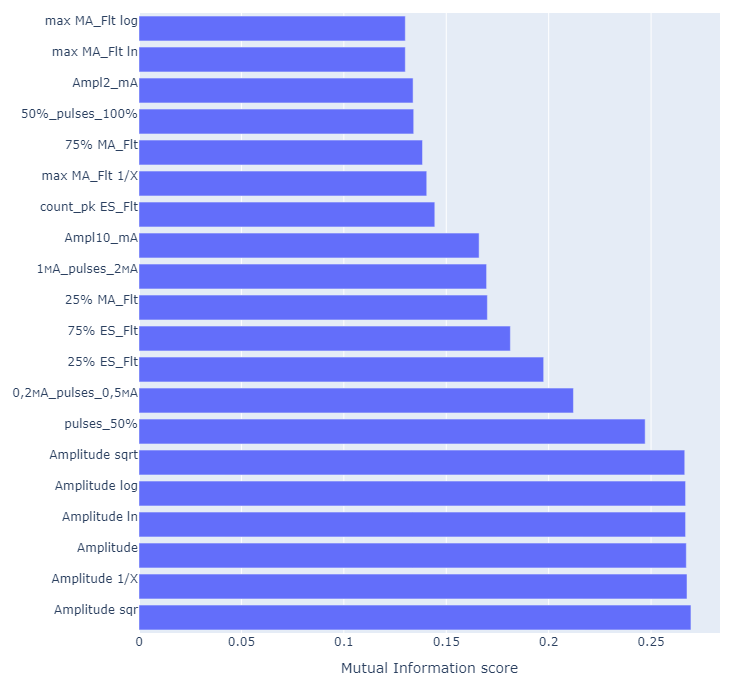}
  \vspace{-3 mm}
  \caption{The top 20 features of the classification task of the proposed method.}
  \label{fig:features_1}
  \end{center}
\end{figure}

The top $20$ features with a mutual information score are presented in Fig.~\ref{fig:features_1}. The amplitude of the fundamental waveform and various functions of this amplitude (square, square root, natural log, common log and inverse proportion) are selected as the most relevant features for the classification task. Moreover, the amplitudes of the $2$nd and $10$th harmonic, pulses in the range of $0.2$ to $0.5$ mA and $1$ to $2$ mA, as well as the pulses less than $100\%$ of fundamental amplitude are accounted for. In addition, some features of the statistical parameters are selected such as the $25$th and $75$th percentile for both MA and ES filters, the number of pulses using the ES filter, and different functions (natural log, common log and inverse proportion) of the absolute maximum between min and max values using MA filter.

\subsection {Validation of the regression task of the proposed method}

In the regression task, the RMSE metric is used to validate both wet and dry models. The RMSE results of various ML models for regression are shown in Table~\ref{table:RMSE}. From these results, it is found that the XGBoost model has the lowest validation error for both wet and dry conditions using the top $10$ features.

\begin{table}[ht]
\centering
 \caption{RMSE of the testing wet and dry regression models for various numbers of selected best features (the regression task).}
 \begin{tabular}{l|l|l|l|l}
 \hline\hline
  Num. of features & LiR & KNN & RF & XGB \\
  (wet model) & [$\%$] & [$\%$]& [$\%$]& [$\%$]\\
 \hline
 1& $6.60$ & $6.35$ & $6.88$ & $7.62$ \\ \hline
 5& $3.37$ & $2.06$ & $1.68$ & $1.06$ \\ \hline
 10 & $2.42$ & $2.06$ & $1.76$ & $\textbf{0.97}$ \\ \hline
 15 & $2.22$ & $2.03$ & $1.66$ & $1.07$ \\ \hline
 20 & $1.94$ & $1.47$ & $1.76$ & $1.21$ \\ \hline
 30 & $1.95$ & $2.95$ & $1.82$ & $1.23$ \\ \hline
 40 & $1.74$ & $12.29$ & $1.87$ & $1.27$ \\ \hline
 64 & $25.97$ & $15.38$ & $2.08$ & $1.45$ \\\hline\hline
 \hline\hline
  Num. of features & LiR & KNN & RF & XGB \\
  (dry model) & [$\%$] & [$\%$]& [$\%$]& [$\%$]\\
 \hline
 1& $1.96$ & $2.15$ & $2.4$ & $2.54$ \\ \hline
 5& $1.93$ & $2.22$ & $2.25$ & $2.34$ \\ \hline
 10 & $1.82$ & $2.09$ & $1.98$ & $\textbf{1.22}$ \\ \hline
 15 & $1.50$ & $2.05$ & $2.01$ & $1.29$ \\ \hline
 20 & $1.46$ & $1.65$ & $2.03$ & $1.32$ \\ \hline
 30 & $1.54$ & $1.93$ & $2.07$ & $1.31$ \\ \hline
 40 & $1.37$ & $4.58$ & $2.08$ & $1.36$ \\ \hline
 64 & $1.41$ & $12.48$ & $2.13$ & $1.40$ \\ \hline\hline
 \end{tabular}
 \label{table:RMSE}
\end{table}

Similar to the classification task, the considered hyperparameters and their optimal values of the XGBoost regression models for the top $10$ input features in the regression task
are presented in Table~\ref{table:RMSE_optim}.

\begin{table}[ht]
\centering
  \caption{Optimal hyperparameters of XGB for regression.}
 \begin{tabular}{l|l|l|l}
 \hline\hline
  Wet model & Hyperparameters & Search range & Optimal value \\
 \hline
\multirow{4}{*}{XGB} 
                    & n-estimators & [$50-1000$] & $732$   \\\cline{2-4}
                    & max depth & [$3-15$] & $7$  \\\cline{2-4}
                    & learning rate & [$0.001-1$] & $0.008$ \\\cline{2-4}
                    & subsample & [$0.5-1$] & $0.5$ \\\cline{2-4}
                    & colsample by tree & [$0.5-1$] & $1.0$ \\ \hline\hline
                
 \hline\hline
  Dry model & Hyperparameters & Search range & Optimal value \\
 \hline
\multirow{4}{*}{XGB} 
                    & n-estimators & [$50-1000$] & $810$   \\\cline{2-4}
                    & max depth & [$3-15$] & $7$  \\\cline{2-4}
                    & learning rate & [$0.001-1$] & $0.016$ \\\cline{2-4}
                    & subsample & [$0.5-1$] & $0.5$ \\\cline{2-4}
                    & colsample by tree & [$0.5-1$] & $1.0$ \\ \hline\hline
                
 \end{tabular}
 \label{table:RMSE_optim}
\end{table}

\begin{figure}[ht]
    \centering
    \subfloat[\centering Wet condition]{{\includegraphics[width=8cm]{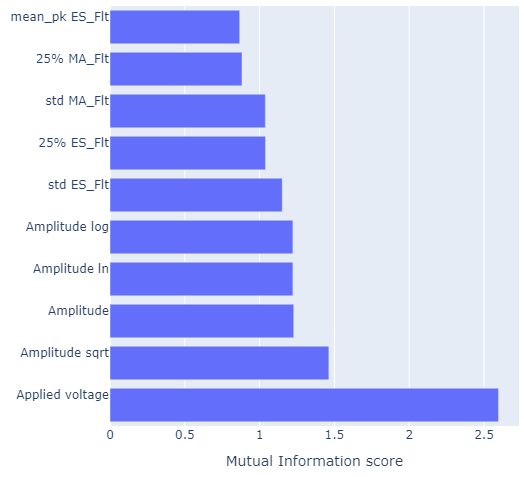} }}%
    \qquad
    \subfloat[\centering Dry condition]{{\includegraphics[width=8cm]{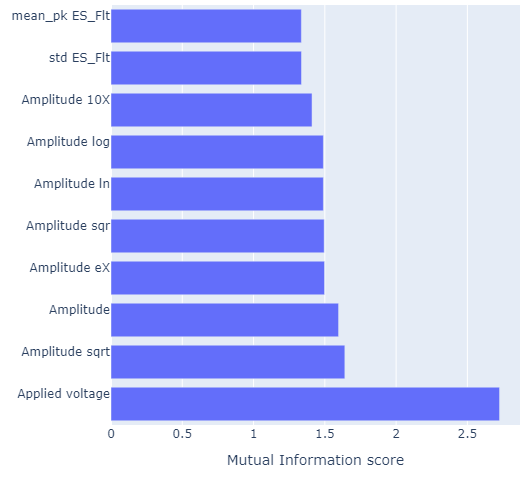} }}%
    \caption{The top $10$ features of the regression task of the proposed method for wet and dry conditions.}%
    \label{fig:features_2}%
    \vspace{-3 mm}
\end{figure}

The applied voltage, amplitude of the fundamental waveform, some functions of this amplitude (square root, natural log, and common log), as well as the standard deviation and mean of all extracted pulses using an ES filter are selected as the most relevant features of the regression task (Fig.~\ref{fig:features_2}). For the wet condition, the $25$th percentile of both MA and ES filters, and the standard deviation using the MA filter are selected as an addition. In the case of the dry condition, the remaining functions of the fundamental waveform amplitude are taken into account such as square, exponential function, inverse proportion, and power of $10$.

\subsection {Validation of the whole proposed method}
The whole method is validated against the rest of the data which is not properly identified in the classification task. The experiments were performed on the XGBoost classification model trained on $10$ and $20$ top features in the classification task. In the regression task, the XGBoost regression models were trained on $1$, $10$, and $30$ top features for both wet and dry cases. The validation results in Table~\ref{table:full} have shown that with a decreasing F1 score for identifying the condition of the insulator, the RMSE error of $U_{50\%}$ voltage estimation increases. However, the increased error is not significant and is within tolerance for the $U_{50\%}$ voltage estimation of the insulator.

\begin{table}[ht]
\caption{RMSE of the full validated method} 
\centering 
\begin{tabular}{c | c | c } 
\hline\hline 
XGB classification & XGB regression & RMSE [$\%$]\\ 
\hline
\multirow{3}{*}{10 features} 
                    & wet - 1 feature & 8.03   \\\cline{2-3}
                    & dry - 1 feature & 3.10  \\\cline{2-3}
                    & wet- 10 features & \textbf{1.51} \\\cline{2-3}
                    & dry - 10 features & \textbf{2.03}
                    \\\cline{2-3}
                    & wet - 30 features & 2.02 \\\cline{2-3}
                    & dry - 30 features & 2.18  \\\hline
                    
\multirow{3}{*}{20 features} 
                    & wet - 1 feature & 7.82   \\\cline{2-3}
                    & dry - 1 feature & 2.89  \\\cline{2-3}
                    & wet- 10 features & \textbf{0.99} \\\cline{2-3}
                    & dry - 10 features & \textbf{1.65} \\\cline{2-3}
                    & wet - 30 features & 1.26 \\\cline{2-3}
                    & dry - 30 features & 1.81  \\\hline
            
\hline\hline 
\end{tabular}
\label{table:full} 
\end{table}

\section{Conclusion}\label{sec:6} 
In this research paper, a ML-based method for outdoor lines
insulator string diagnosis has been proposed. The method uses the XGBoost model to estimate the $U_{50\%}$ voltage of the glass insulator string. Various extracted features from the LC and applied voltage are used as input for the ML models. 

By using the classification model to identify the condition of the insulator (dry or wet) and two separate regression models in the regression task to estimate $U_{50\%}$ voltage, we make the method more robust and accurate. In addition, the results showed that by finding the top relevant features, the $U_{50\%}$ voltage can be estimated with a low error.

The proposed method is implemented in the live insulator monitoring system on $110$kV OHL. Using the applied voltage as an input feature, this method can be extended into every OHL voltage level. In future work, data will be generated from various types different kinds of insulators (ceramic, composite, etc.) and the universal model will be trained to estimate the flashover probability of the insulator strings accurately.


\bibliographystyle{IEEEtran}
\bibliography{sources}

\vfill

\end{document}